\documentclass[runningheads]{llncs}
\usepackage[T1]{fontenc}
\usepackage{graphicx,verbatim}
\usepackage{tabularx}
\usepackage{multirow, booktabs}

\usepackage{kotex}
\usepackage{amssymb}
\usepackage{amsmath}
\usepackage{colortbl}
\usepackage{bm}

\usepackage[table, dvipsnames]{xcolor}

\newcommand{\myparagraph}[1]{\subsubsection*{\textbf{#1}}}
\begin{document}
%
\title{Mono-Modalizing Extremely Heterogeneous Multi-Modal Medical Image Registration}
\titlerunning{Mono-Modalizing Multi-Modal Medical Image Registration}
%

\author{
Kyobin Choo\inst{1} \and
Hyunkyung Han\inst{2} \and
Jinyeong Kim\inst{3} \and
Chanyong Yoon\inst{1} \and\\
Seong Jae Hwang\inst{2}\thanks{Corresponding author}
}

\authorrunning{K. Choo et al.}

\institute{
Department of Computer Science, Yonsei University, Seoul, Republic of Korea \and
Department of Artificial Intelligence, Yonsei University, Seoul, Republic of Korea \and
Yonsei University College of Medicine, Seoul, Republic of Korea \\
\email{\{chu, hhk, jinyeong1324, charlie2019370, seongjae\}@yonsei.ac.kr}}

\maketitle              
\begin{abstract}
In clinical practice, imaging modalities with functional characteristics, such as positron emission tomography (PET) and fractional anisotropy (FA), are often aligned with a structural reference (e.g., MRI, CT) for accurate interpretation or group analysis, necessitating multi-modal deformable image registration (DIR). However, due to the extreme heterogeneity of these modalities compared to standard structural scans, conventional unsupervised DIR methods struggle to learn reliable spatial mappings and often distort images. We find that the similarity metrics guiding these models fail to capture alignment between highly disparate modalities.
To address this, we propose M2M-Reg (Multi-to-Mono Registration), a novel framework that trains multi-modal DIR models using only mono-modal similarity while preserving the established architectural paradigm for seamless integration into existing models. We also introduce GradCyCon, a regularizer that leverages M2M-Reg's cyclic training scheme to promote diffeomorphism. Furthermore, our framework naturally extends to a semi-supervised setting, integrating pre-aligned and unaligned pairs only, without requiring ground-truth transformations or segmentation masks.
Experiments on the Alzheimer’s Disease Neuroimaging Initiative (ADNI) dataset demonstrate that M2M-Reg achieves up to 2× higher DSC than prior methods for PET-MRI and FA-MRI registration, highlighting its effectiveness in handling highly heterogeneous multi-modal DIR.
Our code is available at \url{https://github.com/MICV-yonsei/M2M-Reg}.

\end{abstract}

\section{Introduction}
\label{sec:intro}
Medical imaging encompasses diverse modalities, each capturing distinct physiological and anatomical information. In neuroimaging, modalities such as positron emission tomography (PET) and fractional anisotropy (FA) maps exhibit significant heterogeneity from structural scans due to their specialized purposes.
Thus, practical scenarios often require aligning these scans to a structural reference or spatially normalizing them to a standard MRI space for accurate interpretation and quantitative analysis \cite{fa_reg,pet_reg}. Such cases involve complex, spatially varying deformations, making \textit{multi-modal deformable image registration} (DIR)—which aligns images of different modalities through non-rigid warping—indispensable.

\begin{figure}[t!]
    \centering
    \includegraphics[width=1.0\textwidth]{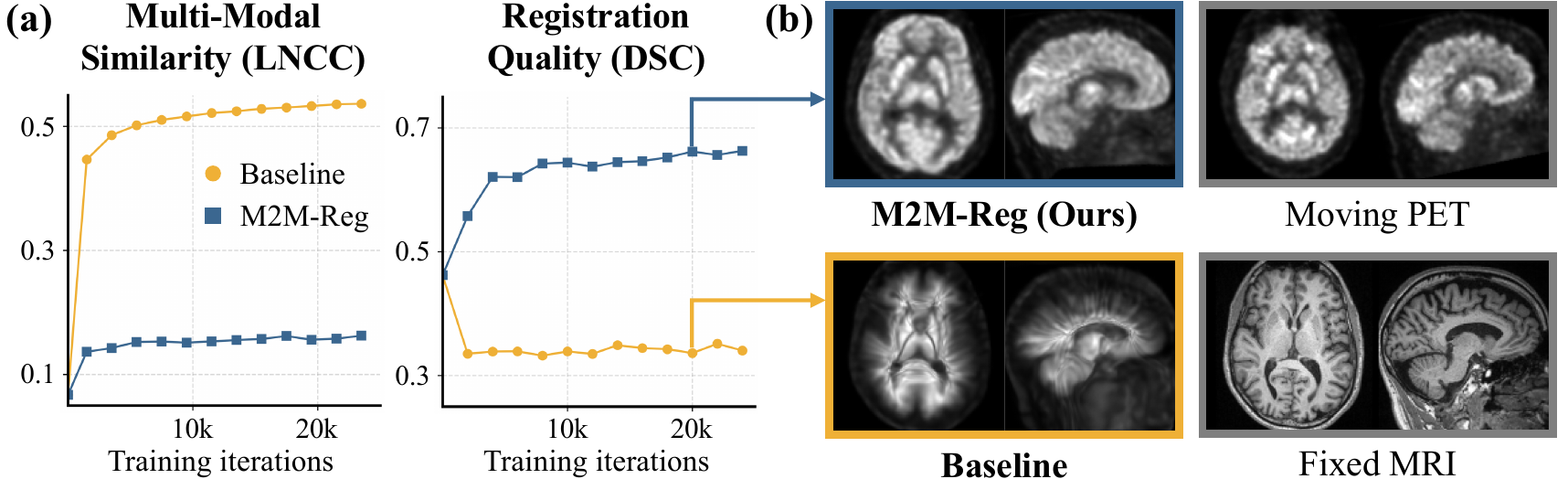}
    \caption{
    \textbf{Discrepancy between image similarity and registration quality in extremely heterogeneous multi-modal image registration.}
    \textbf{(a)} The metric values in the graph show the test set average during training. Our M2M-Reg (blue line) achieves high DSC despite low LNCC, whereas the baseline (yellow line) shows the opposite trend. This suggests that the multi-modal similarity metric does not capture registration quality in highly disparate modalities.
    \textbf{(b)} This discrepant relationship between the two metrics indicates that the baseline fails to establish meaningful PET-MRI correspondences, resulting in severely distorted warped PET images.
    }
    \label{fig:overview}
\end{figure}

In general, multi-modal DIR is performed in an unsupervised manner using unaligned image pairs, since obtaining ground-truth transformations that map the moving image to the fixed image is rarely feasible \cite{reg_survey}. In this setup, the estimated transformation is iteratively refined to maximize the similarity between the fixed and warped moving images. Therefore, the performance of unsupervised DIR hinges heavily on the quality of the similarity metric for measuring image alignment. In multi-modal settings, designing an effective similarity metric is inherently challenging, as images may exhibit different intensity distributions for the same tissue type or even structural variations. To address this, various approaches have been explored, including intensity-based \cite{mmorf,multigradicon}, statistical \cite{mi,mi_used}, self-similarity \cite{mind,ssc}, and deep representation-based \cite{dns,dinoreg} metrics.

While these similarity metrics work well for aligning structural scans such as CT and MRI \cite{multigradicon,dns}, we find that they struggle to establish meaningful spatial correspondences when images exhibit extreme heterogeneity. Concretely, as shown in Fig.~\ref{fig:overview}a, the registration network exhibits counterintuitive behavior in training dynamics: while the average similarity (LNCC) between the warped PET and MRI increases as the training progresses, the registration quality (DSC) becomes even worse than the initial state. This observation indicates that multi-modal similarity metrics do not effectively capture the complex relationship between PET and MRI, especially in regions with intricate and highly detailed anatomical structures, such as the brain (Fig.~\ref{fig:overview}b).

In light of these challenges, we propose \textbf{M2M-Reg}, an innovative framework that enables unsupervised training of multi-modal DIR models using only mono-modal similarity, thereby bypassing the limitations of multi-modal similarity metrics. Our approach trains the model for multiple cross-modal registrations, forming a cyclic source-target relationship.
Specifically, during each training iteration, an additional \textit{bridge pair} is sampled to establish a cyclic source-target mapping in which modalities alternate. By leveraging this cyclic structure, the multi-modal DIR model can be guided by mono-modal similarity as its objective, indirectly learning a more reliable multi-modal mapping.
Additionally, we propose the \textit{Gradient Cycle Consistency} (GradCyCon) regularizer, which encourages diffeomorphism on the warpings within the cycle. GradCyCon is efficiently implemented by penalizing the Jacobian of the cyclic mappings to enforce identity, as these mappings are already computed for mono-modal cycle similarity.

Remarkably, M2M-Reg further enables a new \textit{semi-supervised} DIR paradigm with pre-aligned pairs. By simply designating available pre-aligned pairs as bridge pairs, we can incorporate prior knowledge of the true cross-modal mapping into the model, achieving performance beyond that of unsupervised methods. To the best of our knowledge, this is the first approach to jointly utilize pre-aligned and unaligned pairs within a single framework, without requiring ground-truth transformations or segmentation masks.

\textbf{Contributions.}
Our main contributions are as follows: \textbf{(\romannumeral 1)} We propose M2M-Reg, a novel framework that seamlessly reformulates the multi-modal DIR problem as a mono-modal one while preserving the existing architectural paradigm. \textbf{(\romannumeral 2)} We introduce GradCyCon, a regularization method that naturally leverages the cyclic mapping process in M2M-Reg to promote diffeomorphism. \textbf{(\romannumeral 3)} We further extend M2M-Reg to a pioneering semi-supervised DIR, which encompasses pre-aligned pairs alongside unaligned pairs. \textbf{(\romannumeral 4)} We highlight the limitations of multi-modal similarity in aligning highly heterogeneous modalities and demonstrate that M2M-Reg addresses them, achieving up to 2× higher DSC in brain PET-MRI and FA-MRI registration on ADNI datasets, compared to previous state-of-the-art (SOTA) methods.

\section{Methods}

\subsection{Preliminaries: Unsupervised Learning-Based Paradigm}
\label{sec:method1}
Let source $S: \Omega \to \mathbb{R}$ and target $T: \Omega \to \mathbb{R}$ be grayscale 3D volumes of different modalities, where $\Omega \subset \mathbb{R}^3$ represents the set of spatial coordinates.
A registration network $f_{\theta}$, parameterized by $\theta$, is trained to predict a deformation field $u_{\theta}^{ST}: \Omega \to \mathbb{R}^3$, which represents the displacement vectors mapping coordinates in $T$ to their corresponding locations in $S$.
For arbitrary coordinate $x \in \Omega$, the spatial transformation is then defined as $\Phi_{\theta}^{ST}(x) = x + u_{\theta}^{ST}(x)$, where $\Phi_{\theta}^{ST}: \Omega \to \mathbb{R}^3$ is the transformation function that warps $S$ into alignment with $T$.  
The deformed moving image is obtained as $S \circ \Phi_{\theta}^{ST}$. 
Under this setting, the network parameters $\theta$ are optimized in an unsupervised manner using the following loss function:
\begin{equation}
{\small
\mathcal{L}_\theta = \mathcal{L}_{\text {sim}}(S \circ \Phi_{\theta}^{ST}, T) + \lambda_{\text{reg}} \mathcal{L}_{\text{reg}}(u_{\theta}^{ST}),
}
\label{eq:prelim}
\end{equation}
where $\mathcal{L}_{\text{sim}}(\cdot, \cdot)$ is a multi-modal similarity metric such as local normalized cross-correlation (LNCC) \cite{multigradicon} or mutual information (MI) \cite{ants}, 
$\mathcal{L}_{\text{reg}}(\cdot)$ is a regularization term for enforcing diffeomorphism or promoting smoothness of the deformation field, such as Diffusion or Bending Energy \cite{bending_energy}, and $\lambda_{\text{reg}}$ is a hyperparameter.

\begin{figure}[t!]
    \centering
    \includegraphics[width=1.0\textwidth]{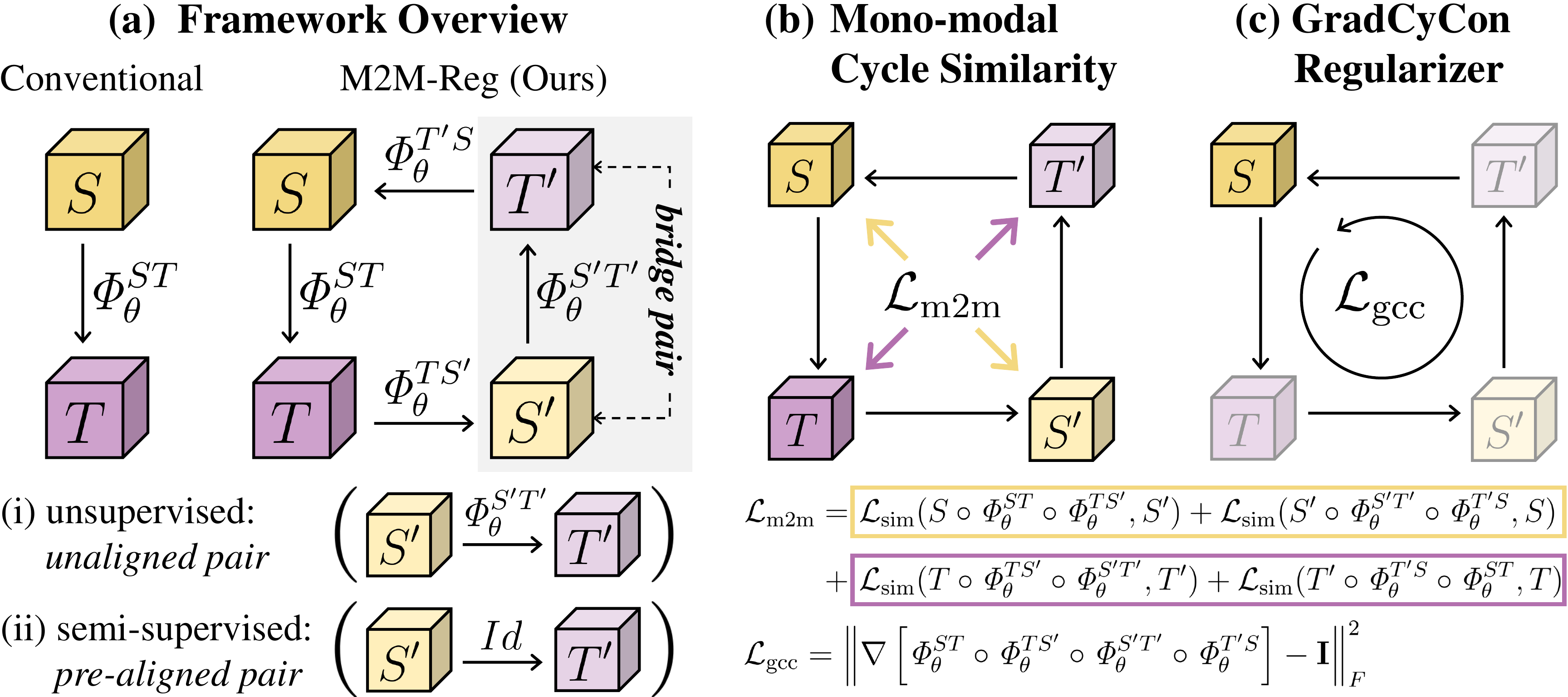}
    \caption{
    \textbf{Overview of the M2M-Reg framework.}
    \textbf{(a)} While the conventional unsupervised multi-modal DIR approach samples a single source-target pair $(S, T)$, M2M-Reg samples an additional \textit{bridge pair} $(S', T')$ to form a cyclic structure. For semi-supervised DIR, the \textit{bridge pair} can be designated as a pre-aligned pair.
    \textbf{(b)} The cyclic structure enables multi-modal DIR models to be guided by a mono-modal similarity objective $\mathcal{L}_{\text{m2m}}$ (Sec.~\ref{sec:method2}).
    \textbf{(c)} For diffeomorphism, the GradCyCon regularizer ($\mathcal{L}_{\text{gcc}}$) encourages the Jacobian ($\nabla$) of the total mapping to approximate identity (Sec.~\ref{sec:method3}). 
    }
    \label{fig:framework}
\end{figure}

\subsection{Learning Multi-Modal Registration via Mono-Modal Similarity}
\label{sec:method2}
The conventional unsupervised multi-modal DIR approach in Sec.~\ref{sec:method1} \cite{multigradicon,coarse_to_fine,transmorph} struggles to establish spatial correspondences in highly heterogeneous modalities (Fig.~\ref{fig:overview}). Here, we introduce an alternative approach that effectively guides multi-modal DIR models using only mono-modal similarity comparisons.

We represent the mappings performed by the multi-modal DIR model $f_\theta$ within a single training iteration as a \textit{graph}. As shown in Fig.~\ref{fig:framework}a (left), the conventional approach samples a single source-target pair $(S, T)$, where $f_\theta$ predicts a directed edge (i.e., mapping) $\Phi_{\theta}^{ST}$ between the two nodes (i.e., images). Since $f_\theta$ is inherently a multi-modal model, it only computes edges between nodes of different modalities. To incorporate mono-modal similarity into the learning process, nodes of the same modality must be connected. The most straightforward approach is to add a reverse edge $\Phi_{\theta}^{TS}$, allowing similarity comparison between $S$ and $S \circ \Phi_{\theta}^{ST} \circ \Phi_{\theta}^{TS}$, as well as between $T$ and $T \circ \Phi_{\theta}^{TS} \circ \Phi_{\theta}^{ST}$. However, this strategy risks the model converging to a trivial solution, such as an identity mapping.

Instead, we introduce a cyclic structure by sampling an additional \textit{bridge pair} $(S', T')$ during each training iteration, adding two nodes and predicting three additional cross-modal edges (Fig.~\ref{fig:framework}a, right). This cyclic design provides several advantages: (\romannumeral 1) It maximizes the number of mono-modal similarity comparisons relative to the number of computed edges, improving efficiency. (\romannumeral 2) The number of similarity comparisons per modality remains consistent across iterations, encouraging stable gradient updates. (\romannumeral 3) The cyclic paths allow composed mappings to approximate identity transformations, naturally enforcing diffeomorphic behavior (Sec.~\ref{sec:method3}).
As a result, two mono-modal paths per modality are formed, as illustrated in Fig.~\ref{fig:framework}b. Leveraging these connections, we formulate an objective purely based on mono-modal similarity, guiding the multi-modal DIR process:
\begin{align}
\small
\mathcal{L}_\text{m2m} = &~ 
\mathcal{L}_{\text{sim}}(S \circ \Phi_{\theta}^{ST} \circ \Phi_{\theta}^{TS'}, S') 
+ \mathcal{L}_{\text{sim}}(S' \circ \Phi_{\theta}^{S'T'} \circ \Phi_{\theta}^{T'S}, S)  \notag \\
+ &~ \mathcal{L}_{\text{sim}}(T \circ \Phi_{\theta}^{TS'} \circ \Phi_{\theta}^{S'T'}, T') 
+ \mathcal{L}_{\text{sim}}(T' \circ \Phi_{\theta}^{T'S} \circ \Phi_{\theta}^{ST}, T),
\label{eq:m2m}
\end{align}
where $\mathcal{L}_{\text{sim}}(\cdot,\cdot)$ can be any mono-modal or multi-modal similarity metric. In our case, we use LNCC, as it has been widely adopted in recent studies \cite{multigradicon,coarse_to_fine,iirp}.

\subsection{Gradient Cycle Consistency}
\label{sec:method3}
The cyclic source-target topology in Sec.~\ref{sec:method2} naturally provides a condition conducive to enforcing diffeomorphic mappings, as the composition of mappings within the cycle should approximate an identity transformation (Fig.~\ref{fig:framework}c). Building on this, inspired by GradICON \cite{gradicon}, we propose the GradCyCon regularizer, which encourages the Jacobian ($\nabla$) of the total mapping to be identity ($\mathbf{I}$):
\begin{equation}
\small
	\mathcal{L}_{\text{gcc}} = \left\|\nabla\left[\Phi_\theta^{ST}\circ\Phi_\theta^{TS'}\circ\Phi_\theta^{S'T'}\circ\Phi_\theta^{T'S}\right]-\mathbf{I} \right\|_F^2.
	\label{eq:gradcycon}
\end{equation}

We now briefly outline the theoretical intuition behind how GradCyCon, like GradICON, implicitly imposes an $H^1$ type regularization effect. Recall that GradICON regularizer enforces diffeomorphism by additionally computing an inverse mapping for a given source-target pair $(S, T)$, as follows:
\begin{equation}
\small
	\mathcal{L}_{\text{reg}}^{\texttt{GradICON}} = \left\|\nabla\left[\Phi_\theta^{ST}\circ\Phi_\theta^{TS}\right]-\mathbf{I} \right\|_F^2.
	\label{eq:gradicon}
\end{equation}
Since the neural network $f_\theta$ inherently produces \textit{independent} noise $n$ with a small $\varepsilon$ (i.e., $\Phi_\theta = \Phi + \varepsilon n$), Eq.~\eqref{eq:gradicon} is approximated as inducing $L^2$ regularization on the inverse of $\nabla\Phi^{ST}$ and $\nabla\Phi^{TS}$, thereby exhibiting an $H^1$-like effect \cite{h1,icon,gradicon}:
\begin{equation}
\small
    \mathbb{E}[\mathcal{L}_{\text{reg}}^{\texttt{GradICON}}]  \approx \varepsilon^2    \Biggl(
    \left\|
    \left[\nabla \Phi^{ST}\right]^{-1} \sqrt{\operatorname{Det}(\nabla\Phi^{ST})}
    \right\|_F^2
    + \left\|
    \left[\nabla\Phi^{TS}\right]^{-1}
    \right\|_F^2\Biggr).
    \label{eq:simplified}
\end{equation}

In GradCyCon (Eq.~\eqref{eq:gradcycon}), grouping three consecutive mappings into one—$\Phi_\theta^{ST}$ and $\hat{\Phi}_\theta^{TS} = \Phi_\theta^{TS'}\circ\Phi_\theta^{S'T'}\circ\Phi_\theta^{T'S}$, or $\hat{\Phi}_\theta^{ST'} = \Phi_\theta^{ST}\circ\Phi_\theta^{TS'}\circ\Phi_\theta^{S'T'}$ and $\Phi_\theta^{T'S}$—yields the same formulation as Eq.~\eqref{eq:gradicon}.
By applying a first-order Taylor approximation, we can express $\hat{\Phi}_\theta^{TS} \approx \Phi^{TS} + \varepsilon \hat{n}^{TS} + O(\varepsilon^2)$. Then, $\hat{n}^{TS}$, formed by three individual ground-truth mappings and their noise components, can be considered independent of $n^{ST}$, and terms of $\varepsilon^2$ and higher are negligible, as assumed in the GradICON. Consequently, Eq.~\eqref{eq:gradcycon} can be derived in the same form as Eq.~\eqref{eq:simplified}, promoting both $\Phi_\theta^{ST}$ and $\hat{\Phi}_\theta^{TS}$, as well as $\hat{\Phi}_\theta^{ST'}$ and $\Phi_\theta^{T'S}$, to be diffeomorphic.

Meanwhile, although the intermediate mappings within the cycle, $\Phi_\theta^{TS'}$ and $\Phi_\theta^{S'T'}$, do not appear to be individually regularized in Eq.~\eqref{eq:gradcycon}, they naturally undergo regularization since target and bridge pairs are randomly resampled during training. From a broader training perspective, GradCyCon eventually enforces diffeomorphism for every mapping. The same holds in the semi-supervised setting (Sec.~\ref{sec:method4}), and we empirically observe that GradCyCon performs effectively.

\subsection{Semi-Supervised Learning with Pre-Aligned Pairs}
\label{sec:method4}
We further extend our framework to incorporate scarce or multiple pre-aligned pairs, allowing the model to reference a reliable multi-modal mapping. The proposed semi-supervised approach enables learning with both unaligned and pre-aligned pairs in an unsupervised manner, offering greater flexibility than conventional supervised methods that rely on ground-truth transformations \cite{gt_sup,gt_sup2} or segmentation masks \cite{seg_sup1,seg_sup2}. As shown at the bottom of Fig.~\ref{fig:framework}a, simply designating the bridge pair $(S', T')$ as a pre-aligned pair naturally integrates prior knowledge of cross-modal correspondence into training. In this setting, since $\Phi^{S'T'}$ can be regarded as the identity mapping, we omit $\Phi_\theta^{S'T'}$ from the formulations of our mono-modal cycle similarity (Eq.~\eqref{eq:m2m}) and GradCyCon regularizer (Eq.~\eqref{eq:gradcycon}), thereby simplifying optimization. As a result, one fewer mapping needs to be computed, enhancing efficiency while further improving performance.

\begin{figure}[t!]
    \centering
    \includegraphics[width=1.0\textwidth]{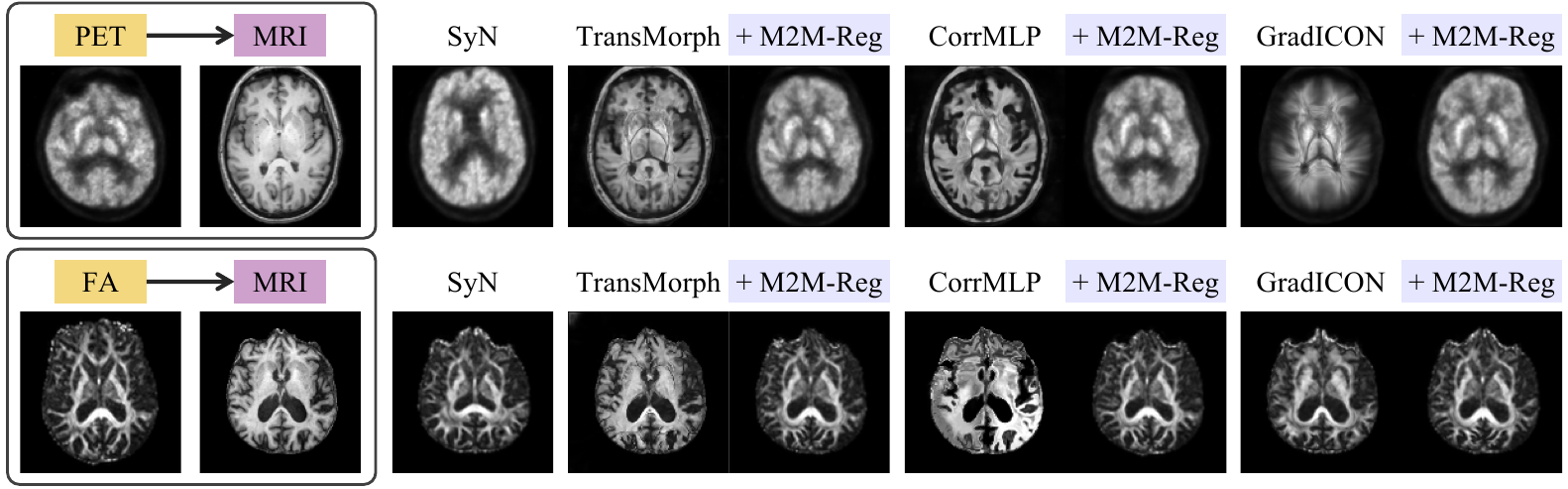}
    \caption{\textbf{Qualitative comparison of PET-MRI and FA-MRI registration}}

    \label{fig:qual_res}
\end{figure}

\subsection{Final Objective: Integration with Existing Models}
\label{sec:method5}
The proposed M2M-Reg is a flexible framework that can be seamlessly applied to any unsupervised approach that follows a formulation similar to Eq.~\eqref{eq:prelim}. It simply replaces the conventional multi-modal similarity with our mono-modal cycle similarity and incorporates GradCyCon, leading to the following final objective:
\begin{equation}
{\small
\mathcal{L}_\theta^{\text{final}} = \mathcal{L}_{\text{m2m}} + \lambda_{\text{reg}} \mathcal{L}_{\text{reg}} + \lambda_{\text{gcc}} \mathcal{L}_{\text{gcc}},
}
\label{eq:final}
\end{equation}
where $\mathcal{L}_{\text{reg}}$ is the original regularization, and $\lambda_{\text{reg}}$ and $\lambda_{\text{gcc}}$ are hyperparameters.

\section{Experiments}
\label{sec:experiments}

\begin{table}[t]
	\centering
	\caption{\textbf{Quantitative comparison of PET-MRI and FA-MRI registration}}
	\label{tab:result}
        \resizebox{\textwidth}{!}{
		\begin{tabular}{lcccccccc}
            \specialrule{1pt}{0pt}{3pt}
			\multirow{2.4}{*}{Method} & \multirow{2.4}{*}{$\mathcal{L}_{reg}$} & \multirow{2.4}{*}{$\mathcal{L}_{sim}$} & \multicolumn{2}{c}{M2M-Reg} & \multicolumn{2}{c}{PET $\rightarrow$ MRI} & \multicolumn{2}{c}{FA $\rightarrow$ MRI} \\
			\cmidrule(lr){4-5} \cmidrule(lr){6-7} \cmidrule(lr){8-9}
			& & & \rule{1pt}{0ex} $\mathcal{L}_{\text{m2m}}$ & $\mathcal{L}_{\text{gcc}}$ \rule{1pt}{0ex} & \rule{1pt}{0ex} DSC $\uparrow$ & $\%|J_\phi|_{<0}\downarrow$ \rule{1pt}{0ex} & \rule{1pt}{0ex} DSC $\uparrow$ & $\%|J_\phi|_{<0}\downarrow$ \rule{1pt}{0ex} \\
            \specialrule{0.5pt}{2.5pt}{0pt}
			Initial & - & - & \rule{1pt}{0ex} - & - \rule{1pt}{0ex} & \rule{1pt}{0ex} 0.462 & - \rule{1pt}{0ex} & \rule{1pt}{0ex} 0.424 & - \rule{1pt}{0ex} \\
            
            \specialrule{0.5pt}{0pt}{0pt}
			SyN \cite{ants} & Gaussian & MI & \rule{1pt}{0ex} - & - \rule{1pt}{0ex} & \rule{1pt}{0ex} 0.262 & 0.000 \rule{1pt}{0ex} & \rule{1pt}{0ex} 0.635 & 0.000 \rule{1pt}{0ex} \\
            NiftyReg \cite{niftyreg} & BE & LNCC & \rule{1pt}{0ex} - & - \rule{1pt}{0ex} & \rule{1pt}{0ex} 0.196 & 0.000 \rule{1pt}{0ex} & \rule{1pt}{0ex} 0.607 & 0.038 \rule{1pt}{0ex} \\
            FireANTs \cite{fireants} & Gaussian & MI & \rule{1pt}{0ex} - & - \rule{1pt}{0ex} & \rule{1pt}{0ex} 0.223 & 1.737 \rule{1pt}{0ex} & \rule{1pt}{0ex} 0.420 & 0.094 \rule{1pt}{0ex} \\
            
            \specialrule{0.5pt}{0pt}{0pt}
            \multirow{3}{*}{TransMorph \cite{transmorph}} & \multirow{3}{*}{Diffusion} & \multirow{3}{*}{LNCC} & \rule{1pt}{0ex}  &  \rule{1pt}{0ex} & \rule{1pt}{0ex} 0.322 & 40.572 \rule{1pt}{0ex} & \rule{1pt}{0ex} 0.449 & 39.534 \rule{1pt}{0ex} \\
            & & & \rule{1pt}{0ex} \checkmark &  \rule{1pt}{0ex} & \rule{1pt}{0ex} 0.496 & 8.020 \rule{1pt}{0ex} & \rule{1pt}{0ex} 0.510 & 10.464 \rule{1pt}{0ex} \\
            & & & \rule{1pt}{0ex} \cellcolor{blue!10}\checkmark & \cellcolor{blue!10}\checkmark \rule{4pt}{0ex} & \cellcolor{blue!10}\rule{1pt}{0ex} 0.657 & \cellcolor{blue!10}0.003 \rule{1pt}{0ex} & \cellcolor{blue!10}\rule{1pt}{0ex} 0.641 & \cellcolor{blue!10}0.000 \rule{1pt}{0ex} \\
            
            \specialrule{0.5pt}{0pt}{0pt}
            \multirow{3}{*}{CorrMLP \cite{coarse_to_fine}} & \multirow{3}{*}{Diffusion} & \multirow{3}{*}{LNCC} & \rule{1pt}{0ex}  &  \rule{1pt}{0ex} & \rule{1pt}{0ex} 0.237 & 43.149 \rule{1pt}{0ex} & \rule{1pt}{0ex} 0.064 & 50.106 \rule{1pt}{0ex} \\
            & & & \rule{1pt}{0ex} \checkmark &  \rule{1pt}{0ex} & \rule{1pt}{0ex} 0.583 & 0.422 \rule{1pt}{0ex} & \rule{1pt}{0ex} 0.600 & 0.003 \rule{1pt}{0ex} \\
            & & & \rule{1pt}{0ex} \cellcolor{blue!10}\checkmark & \cellcolor{blue!10}\checkmark \rule{4pt}{0ex} & \cellcolor{blue!10}\rule{1pt}{0ex} 0.587 & \cellcolor{blue!10}0.000 \rule{1pt}{0ex} & \cellcolor{blue!10}\rule{1pt}{0ex} 0.607 & \cellcolor{blue!10}0.000 \rule{1pt}{0ex} \\
            
            \specialrule{0.5pt}{0pt}{1pt}
            \multirow{4}{*}{GradICON \cite{gradicon}} & \multirow{4}{*}{GradICON} & $\text{LNCC}^2$ \cite{multigradicon} & \rule{1pt}{0ex}  &  \rule{1pt}{0ex} & \rule{1pt}{0ex} 0.088 & 5.508 \rule{1pt}{0ex} & \rule{1pt}{0ex} 0.632 & 0.063 \rule{1pt}{0ex} \\
            \cline{3-9}
            & & \multirow{3}{*}{LNCC} & \rule{1pt}{0ex}  &  \rule{1pt}{0ex} & \rule{1pt}{0ex} 0.346 & 0.041 \rule{1pt}{0ex} & \rule{1pt}{0ex} 0.647 & 0.008 \rule{1pt}{0ex} \\
            & & & \rule{1pt}{0ex} \checkmark &  \rule{1pt}{0ex} & \rule{1pt}{0ex} 0.668 & 0.000 \rule{1pt}{0ex} & \rule{1pt}{0ex} 0.662 & 0.000 \rule{1pt}{0ex} \\
            & & & \rule{1pt}{0ex} \cellcolor{blue!10}\checkmark & \cellcolor{blue!10}\checkmark \rule{4pt}{0ex} & \cellcolor{blue!10}\rule{1pt}{0ex} 0.676 & \cellcolor{blue!10}0.000 \rule{1pt}{0ex} & \rule{1pt}{0ex} \cellcolor{blue!10}0.666 & \cellcolor{blue!10}0.000 \rule{1pt}{0ex} \\
            \specialrule{1pt}{0pt}{0pt}
		\end{tabular}
	}
\end{table}

\myparagraph{Datasets.}
We constructed paired PET-MRI (train: 300, test: 50) and FA-MRI (train: 230, test: 50) datasets using brain 18F-Fluorodeoxyglucose PET, FA, and T1-weighted MRI from the ADNI\footnote{\url{http://adni.loni.usc.edu}} \cite{adni,adni2} study. All data were resampled to 1.5 mm isotropic voxels with a 128×128×128 resolution, and the FA-MRI set was skull-stripped using SynthStrip \cite{synthstrip}. Intensities were clipped at the upper 99.9th percentile and min-max normalized to [0,1]. Each pair was rigidly coregistered using SPM12 \cite{spm} for alignment. For DSC evaluation, MRI was segmented using FreeSurfer v7.1.1 \cite{freesurfer}, with the resulting labels grouped into 11 classes.

\myparagraph{Comparison Methods.}
We compare optimization-based methods (SyN \cite{ants}, NiftyReg \cite{niftyreg}, FireANTs \cite{fireants}) and unsupervised learning models, including TransMorph \cite{transmorph} (a widely used transformer-based model), CorrMLP \cite{coarse_to_fine} (a SOTA MLP-based architecture), and GradICON \cite{gradicon} (a CNN-based model with a SOTA regularizer), evaluating their performance with and without M2M-Reg.

\myparagraph{Implementation Details.}
M2M-Reg was implemented on the GradICON codebase and trained with an NVIDIA RTX A6000 48GB GPU using ADAM optimizer with a learning rate of 5e-5 and batch size of 8, for 50,000 iterations. The regularization weights were set to $\lambda_\text{reg} = 0.5$ (1.5 without M2M-Reg) and $\lambda_\text{gcc} = 0.1$. For TransMorph and CorrMLP, a learning rate of 5e-5, batch size of 2 were used for 50,000 iterations, with regularization weights of $\lambda_\text{reg} = 0.01$ (1000 without M2M-Reg) and $\lambda_\text{gcc} = 1$. Hyperparameters were carefully optimized across all baselines. In all methods, a single network $f_\theta$ was used to perform bidirectional mappings between the two modalities. Throughout training, source-target pairs were always drawn from different subjects, except for bridge pairs in the semi-supervised setting.

\subsection{Evaluations}
\label{sec:exp2}

\myparagraph{Quantitative Results.}
We evaluated registration quality using the Dice similarity coefficient (DSC) and deformation quality via the percentage of negative Jacobian determinants ($\%|J_\phi|_{<0}$). In Table~\ref{tab:result}, most baselines exhibited severe $\%|J_\phi|_{<0}$ and showed little to no DSC improvement, even performing worse than the initial state in PET$\rightarrow$MRI. Applying M2M-Reg to baselines (highlighted in blue) significantly improved DSC (up to 2×) while nearly eliminating $\%|J_\phi|_{<0}$, demonstrating our framework's effectiveness in handling high heterogeneity. Even using only mono-modal cycle similarity ($\mathcal{L}_{\text{m2m}}$) led to substantial improvements, confirming its role in guiding the model to learn meaningful mappings. GradCyCon ($\mathcal{L}_{\text{gcc}}$) further enhanced both metrics by preventing unrealistic deformations. Notably, its effect was less pronounced in GradICON, likely because GradCyCon performed a similar implicit $H^1$-type regularization.

\begin{table}[t]
    \caption{\textbf{Performance across pre-aligned bridge pair ratios (PET $\rightarrow$ MRI)}}
    \centering
    \renewcommand{\arraystretch}{1.4}
    \renewcommand{\tabcolsep}{4pt} 
    \begin{tabular}{l|ccccccc}
        \specialrule{1pt}{0pt}{0pt}
        \# Pre-Aligned Pairs \rule{1pt}{0ex} & 0 (unsup.) & 1\% (3 pairs) & 5\% & 10\% & 25\% & 50\% & 100\% \\ 
        \specialrule{0.5pt}{0pt}{0pt}
        DSC $\uparrow$ \rule{1pt}{0ex} & 0.676 & 0.691 & 0.709 & 0.709 & 0.725 & 0.729 & 0.730 \\ 
        $\%|J_\phi|_{<0}\downarrow$ \rule{1pt}{0ex} & 0.000 & 0.000 & 0.000 & 0.000 & 0.000 & 0.009 & 0.000 \\ 
        \specialrule{1pt}{0pt}{0pt}
    \end{tabular}
    \label{tab:semi_sup}
\end{table}

\myparagraph{Qualitative Results.}
In Fig.~\ref{fig:qual_res}, baselines struggled to establish spatial mapping between disparate modalities. In PET-MRI registration, they either forced PET into MRI-like shapes (TransMorph, CorrMLP) or severely collapsed its structures (GradICON). In FA-MRI registration, they artificially expanded FA’s white matter to resemble cortical regions in MRI. In both cases, SyN failed to deform regions like the ventricles sufficiently. In contrast, M2M-Reg accurately aligned to the target anatomy while preserving the source modality’s structures.

\myparagraph{Semi-Supervised Setting.}
We evaluated semi-supervised DIR by gradually increasing the proportion of pre-aligned bridge pairs in M2M-Reg within GradICON. In Table~\ref{tab:semi_sup}, performance improved monotonically, with DSC increasing by up to 5.4\%p, proving its significant impact. Notably, even using just 3 pairs improved DSC by 1.5\%p, and employing 25\% of pairs achieved performance comparable to 100\%, highlighting the effectiveness of the proposed framework.

\section{Conclusion}
\label{sec:conclusions}

Recognizing that multi-modal similarity is inadequate for registering highly heterogeneous modalities, we introduced M2M-Reg, a paradigm-shifting framework that reformulates multi-modal deformable image registration (DIR) as a mono-modal problem. By leveraging only mono-modal similarity within a novel cyclic learning scheme, M2M-Reg circumvented the limitations of conventional unsupervised DIR, which relies on multi-modal similarity, while preserving architectural flexibility. Coupled with GradCyCon, which approximates diffeomorphic transformations through cycle consistency, our framework effectively captures complex heterogeneous mappings in an unsupervised manner.
Empirical results demonstrated that M2M-Reg consistently improved performance across diverse baselines and datasets, successfully handling highly disparate modalities. Beyond its strong unsupervised performance, our semi-supervised approach further maximizes the utility of pre-aligned pairs. This adaptability positions M2M-Reg as a practical, scalable solution for advancing multi-modal medical image analysis.

\subsubsection{\ackname} This work was supported in part by the IITP RS-2024-00457882 (AI Research Hub Project), IITP 2020-II201361, NRF RS-2024-00345806, NRF RS-2023-00219019, and NRF RS-2023-002620.

\subsubsection{Disclosure of Interests.} The authors have no competing interests.

%
%
%
\bibliographystyle{splncs04}
\bibliography{main}
\end{document}